\documentclass[runningheads]{llncs}
\usepackage{float}
\usepackage{pifont}
\usepackage{footnote}
\usepackage{enumitem}
\usepackage{bm}
\usepackage{arydshln}
\usepackage{booktabs}
\usepackage{multicol}
\usepackage{multirow}
\usepackage{color}
\usepackage{xcolor}     
\usepackage{colortbl}
\usepackage{soul}
\usepackage{bbding}
\usepackage{makecell}
\usepackage{mathtools}
\usepackage{imakeidx}
\usepackage{amssymb}
\usepackage{graphicx}
\usepackage{amsmath}
\usepackage{threeparttable}
\definecolor{citecolor}{HTML}{0071BC}
\definecolor{linkcolor}{HTML}{ED1C24}
\usepackage{hyperref}
\makeindex
\usepackage{arydshln}
\usepackage{lipsum}
\usepackage[toc]{multitoc}
\usepackage[edges]{forest}
\usepackage[normalem]{ulem}

\usepackage{bbding}
\usepackage[most]{tcolorbox}

\usepackage{algorithm}
\usepackage{algorithmic}

\usepackage{minitoc}
\usepackage[toc,page,header]{appendix}

\definecolor{orchid}{rgb}{0.85, 0.44, 0.84}
\definecolor{rubinred}{rgb}{0.82, 0.0, 0.28}
\definecolor{flagship}{rgb}{0.93, 0.06, 0.41}
\definecolor{radiologist}{rgb}{0.50, 0.50, 1}
\definecolor{YT}{HTML}{002FA7}

\newcolumntype{P}[1]{>{\centering\arraybackslash}p{#1}}
\newlength\savewidth
  
\usepackage{xspace}
\newcommand{\eyepose}{{\fontfamily{ppl}\selectfont GenEyePose}\xspace}
\usepackage[T1]{fontenc}
\usepackage{graphicx,verbatim}
\usepackage{color}

\usepackage{tabularx}
\usepackage[misc]{ifsym}
\newcommand{\corrAuthor}{$^{\textrm{\Letter}}$}

\begin{document}

\title{\eyepose: Patient-Free, Knowledge-Based Saccadic Eye Movement Modeling for Digital Neurophysiologic Biomarker Development}
\titlerunning{\eyepose}
%
\author{
Tianyu Lin\inst{1} \and 
Jooyoung Ryu\inst{1} \and  
Puvada Sreevarsha\inst{1} \and    
Rahul Srinivasaragavan\inst{1} \and    
Riya Satavlekar\inst{1} \and    
Susan Kim\inst{1} \and    
Nidhi Soley\inst{1} \and    
Yujie Yan\inst{1} \and  
Ishan Vatsaraj\inst{1} \and 
Carl Harris\inst{1} \and    
Aimon Rahman\inst{1} \and   
Vishal Patel\inst{1} \and   
Joseph Greenstein\inst{1} \and  
Casey Taylor\inst{1}\and    
Kemar E. Green\inst{1,2,3} \corrAuthor    
}
\authorrunning{T. Lin et al.}

\institute{Whiting School of Engineering, Johns Hopkins University, Baltimore, USA \and
Johns Hopkins Data Science and AI Institute, Baltimore, USA \and
NeuroAgent AI, Inc., Baltimore, USA \\
\email{kemarearlgreen@neuroagentai.org}\\}


\maketitle              

\begin{abstract}
Eye movements, including saccades, are widely regarded as highly sensitive and objective biomarkers of neurophysiologic states. Detecting saccadic signatures in neurologic diseases offers a rapid, portable alternative to brain imaging, avoiding access and cost barriers. Currently, there are no robust AI-enabled video-oculographic solutions (e.g., digital biomarkers) for screening, triaging, or localizing brain abnormalities due to privacy issues and scarce datasets. In this work, we propose the first fully synthetic, patient-free, multimodal eye movement generation pipeline for generalizable saccade analysis. Using this synthetic dataset, we trained a deep learning classifier to distinguish between normal and abnormal (hypometria and hypermetria) saccadic accuracies and evaluated its performance on real-world clinical data. 
The model achieved an AUROC of 0.76 and a sensitivity of 0.71, showing that the synthetic data has strong potential to generalize for clinical applications, including as a screening tool in at-home and emergency room settings or a tool for precise neuroanatomic localization. 
\end{abstract}
\begin{figure}[htb]
    \centering
    \includegraphics[width=\textwidth]{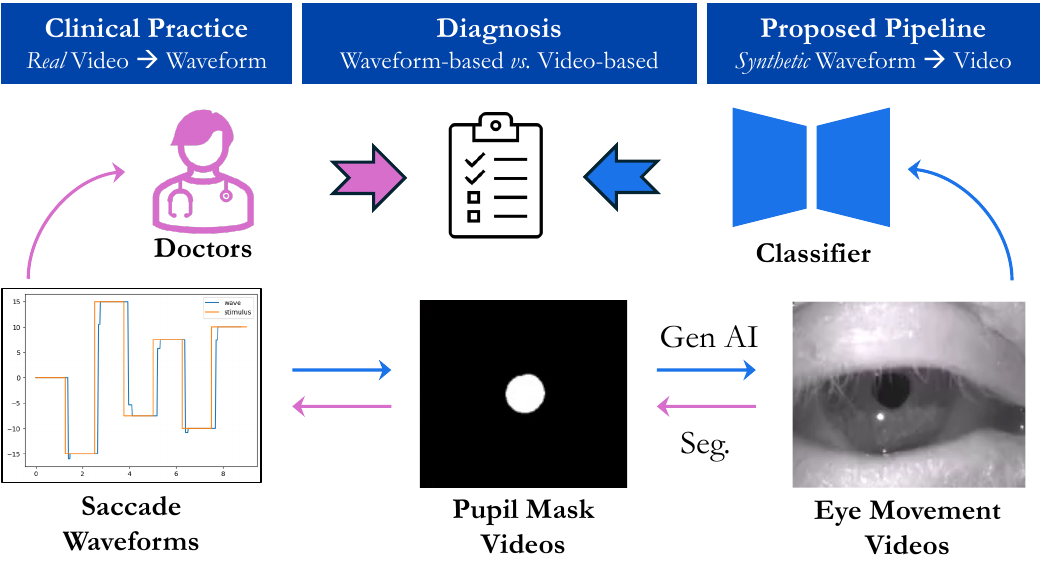}
    \caption{\textbf{\eyepose framework.} \textcolor{magenta}{Clinical Practice}: Pupils are extracted from real patient eye videos to generate waveforms for diagnosis. \textcolor{blue}{Proposed Pipeline}: Synthetic eye movement videos are generated from synthetic waveforms to train deep classifiers.}
    \label{figs:overview}
\end{figure}

\section{Introduction}

Neurological disorders affect over 3 billion individuals globally~\cite{steinmetz2024global}. Current diagnostic tools like magnetic resonance imaging (MRI) and computed tomography (CT) are often costly, inaccessible, and time-consuming. Alternatively, eye movements, particularly saccades, offer a highly sensitive, objective, and portable biomarker for localizing neurological diseases~\cite{leigh2015neurology,kocak2021novel,kattah2022video}.Saccades are rapid ballistic eye movements that shift gaze between targets at velocities as high as 700°/s. These movements are initiated and controlled by complex neural pathways in brain regions spanning the cortex, subcortex, brainstem, and cerebellum \cite{leigh2015neurology}. Saccadic metrics such as latency, accuracy, and velocity exhibit distinct patterns across different neurological states~\cite{leigh2015neurology}.

However, robust machine learning models for saccadic analysis remain underdeveloped~\cite{Przybyszewski2023-tt}. The primary bottleneck is the lack of large, annotated video-oculographic datasets, which are severely limited by patient privacy concerns and biometric data restrictions~\cite{graham2024digital,6117536}. Furthermore, real-world eye movement data is frequently degraded by noise, blinks, and low frame rates, complicating reliable feature extraction and model development~\cite{wagle2022-46,mukunda2024deep}.

To address data scarcity, prior works have explored synthetic data generation using autoencoders, convolutional neural networks, and diffusion models to augment existing medical datasets~\cite{jiao2024diffeyesyn,rahman2025generative,Rahman2026GenVOG,Rahman2026GenVOGDIT}. Yet, the use of fully synthetic data as the exclusive source for model training remains largely unexplored in this domain~\cite{wang2023deep}. Building on computational models of saccadic dynamics~\cite{10.1093/brain/awl309}, we leverage generative AI to produce knowledge-based physiologically plausible synthetic data without relying on real patient records.

In this work, we propose a fully synthetic, patient-free, multimodal eye movement generation pipeline. We generate synthetic saccade waveforms using clinically derived parameters, translate them into pose-guided pupil mask videos, and generate realistic eye movement videos via a conditional diffusion model~\cite{zhang2023adding,mei2022vidm}. We then train deep learning classifiers on this synthetic data and evaluate their generalizability on real-world clinical data.

Our main contributions are as follows:
\begin{itemize}
    \item We propose the first fully synthetic, patient-free, multimodal video generation pipeline for generalizable saccade analysis.
    \item We incorporate clinically established physiological parameters to ensure the synthetic waveforms and videos maintain high physiological plausibility.
    \item We demonstrate that deep learning models trained exclusively on our synthetic dataset can generalize to unseen, real-world clinical data, serving as a potential digital neurophysiologic biomarker.
\end{itemize}

\section{Methods}

\eyepose first generates synthetic saccadic waveforms that closely mirror true physiological dynamics. We then create pupil mask videos by mapping these waveforms to spatial movements. Finally, a video generation model conditioned on these masks produces realistic eye movement videos. We utilize this synthetic dataset to train a deep learning classifier to differentiate normal, hypometric, and hypermetric saccades, evaluating its generalizability on real-world clinical data. Figure~\ref{figs:overview} illustrates this workflow.

\subsection{Waveform Generation}
To develop synthetic waveforms modeling saccades, we created an algorithm to generate 60-Hz monocular waveforms representing normal, bilateral hypometric, and bilateral hypermetric saccades. 

A fundamental property that all saccades follow is the main sequence: as the amplitude of a saccadic shift increases, the peak velocity also increases~\cite{BehaviorResearchMethods}. Additionally, bilateral hypermetric and hypometric saccades exhibit higher latencies and lower precision compared to normal saccades. We incorporated clinically accepted parameter ranges into our algorithm~\cite{bahill1975dynamic}: normal saccade accuracy ranges from 70–120\%, hypermetric from 120–150\%, and hypometric from 10–70\%. Normal saccades have a latency of 0.0–0.4 seconds, whereas both abnormal classes range from 0.4–0.7 seconds.

Based on prior studies analyzing age-related parameter distributions~\cite{Age}, we sampled these metrics from normal distributions. Normal saccade accuracy was sampled from $\mathcal{N}(95, 10^2)$, hypermetric from $\mathcal{N}(135, 5^2)$, and hypometric from $\mathcal{N}(40, 10^2)$. Latency was sampled from $\mathcal{N}(0.3, 0.025^2)$ for normal saccades and $\mathcal{N}(0.55, 0.05^2)$ for abnormal classes.

To construct the waveforms, we generate a target sequence representing roughly 5 seconds of a video-oculography response (VOR) test. Target amplitudes are randomly selected from $\pm\{2.5, 7.5, 10, 15\}$ degrees. Every waveform follows the main sequence and its assigned clinical parameters. Finally, random Gaussian noise ($\mathcal{N}(0, 1)$) is added to simulate natural patient artifacts. These fully annotated waveforms subsequently guide the pupil mask videos.

\begin{figure}[ht]
    \centering
    \includegraphics[width=\linewidth]{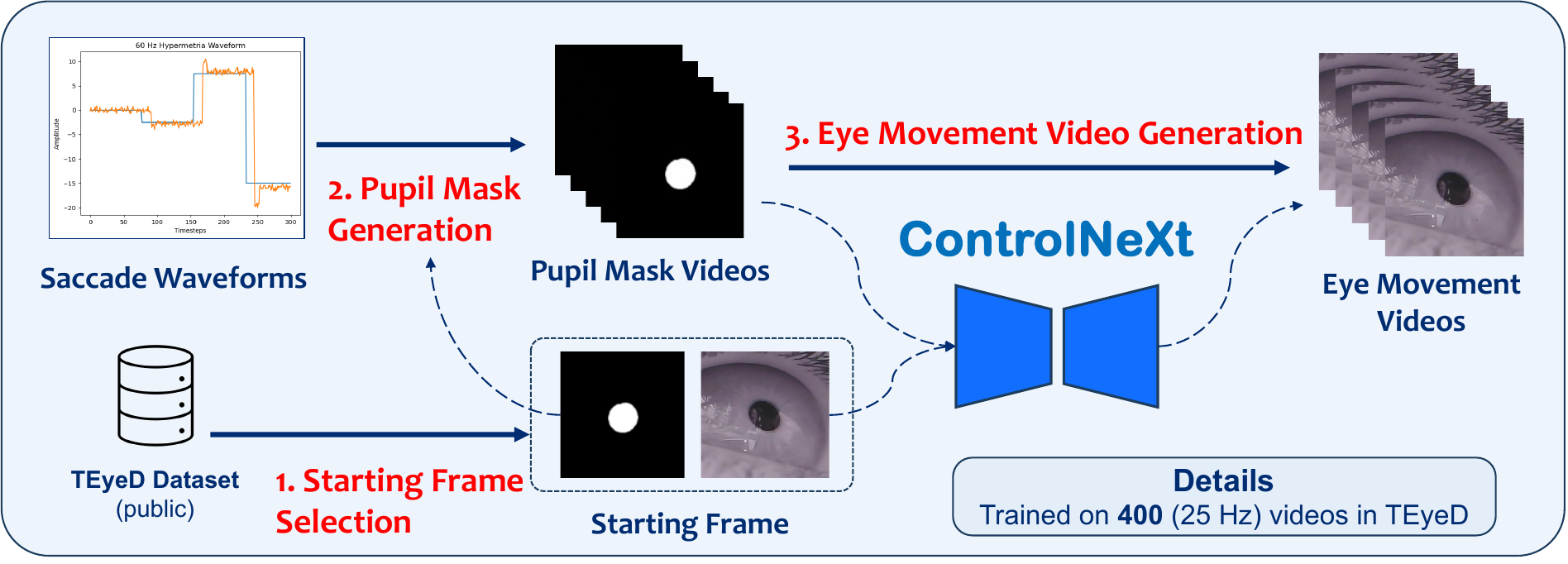}
    \caption{\textbf{The overview of the proposed eye movement video generation pipeline.} Note that only a publicly available eye video dataset is used in the pipeline, and no real patient data is needed. }
    \label{fig:video-gen}
\end{figure}

\subsection{Video Generation Pipeline}
We propose a pose-guided video generation pipeline that utilizes the generated waveforms to synthesize corresponding eye movement videos~\cite{rahman2025generative}. While traditional eye tracking extracts pupil masks to generate waveforms~\cite{steinhart2023video,santini2018pure,de2006optimal}, our pipeline reverses this process: we map synthetic waveforms to pupil mask videos, which then condition a generative model to produce realistic eye videos. The pipeline overview is shown in Fig.~\ref{fig:video-gen}.

The pipeline begins by selecting a starting frame from the public TEyeD dataset~\cite{Fuhl2021TEyeDO2}. We randomly choose an image and its corresponding pupil mask, discarding frames with empty masks (e.g., during blinks). This guarantees high initial frame quality and provides a robust visual anchor for the pose-guided generation model.

To generate the sequential pupil mask video, we model horizontal saccadic amplitudes. We create synthetic pupil masks represented as ellipses with randomized eccentricities, orientations, and aspect ratios to enhance morphological diversity. Assuming the pupil shape remains constant during movement, we establish a simple linear mapping between the waveform's amplitude and the pupil mask's horizontal position within the frame. This yields a frame-by-frame pupil mask video synchronized to the waveform.

We selected ControlNeXt~\cite{peng2024controlnext} as our video generation engine. Pretrained on natural videos, this model utilizes mask conditions to guide structural movements. We further fine-tuned ControlNeXt on the TEyeD dataset using pairs of pupil mask videos and eye movement videos. 

By inputting the starting frame and the synthesized pupil mask video, our model generates a realistic eye movement video that inherently shares the diagnostic label of the driving waveform. While real clinical data frequently contains complex artifacts (e.g., square wave jerks) that complicate labeling, our synthetic waveforms strictly adhere to the predefined clinical patterns of the three saccade classes, yielding a rigorously controlled training dataset.

\section{Experimental Setup}

\subsection{Datasets}
\subsubsection{Synthetic Dataset.} The synthetic dataset was generated using clinically established parameter distributions. The sampled accuracy was $\mathcal{N}(95, 10^2)$ for normal saccades, $\mathcal{N}(40, 10^2)$ for hypometric saccades, and $\mathcal{N}(135, 5^2)$ for hypermetric saccades~\cite{Age}. These waveforms strictly follow physiological main sequence principles, ensuring realistic amplitude and velocity relationships. The generated data were used exclusively for model training and in-distribution validation.

\subsubsection{Clinical Dataset.} We evaluated our method on a private clinical dataset of 113 patients with various neurovestibular disorders, acquired and annotated by a single eye-movement clinical expert at The Johns Hopkins Hospital (50 normal, 15 with hypermetria, and 48 with hypometria). Derived from 15-second horizontal stimulus video-oculography tests, the dataset contains inherent real-world variability and overlapping eye movement patterns. This dataset was used exclusively to evaluate the out-of-distribution generalizability of our trained models.

\subsection{Implementation Details}
The video generation model, ControlNeXt~\cite{peng2024controlnext}, was fine-tuned on the public TEyeD dataset~\cite{Fuhl2021TEyeDO2} using two NVIDIA A800 GPUs. For saccade classification, we implemented the MViT-V2 model~\cite{fan2021multiscale} using Torchvision~\cite{marcel2010torchvision}, trained with an Adam optimizer at a learning rate of 0.0001 for 40 epochs.

\subsubsection{Clinical Labeling Protocol.} Ground truth labels for the clinical dataset were established by evaluating saccadic amplitude accuracy and validated by one clinical eye movement expert. Cases exhibiting two or more consecutive abnormal saccades on either side were labeled as hypermetric or hypometric. All flagged cases were independently reviewed by a second annotator, and only those with matching labels were retained, ensuring high confidence in the evaluation set.

\subsubsection{Model Training and Inference.} During training, we addressed the sparsity of saccadic movements by randomly sampling short video clips aligned with the stimulus onset, adding a randomized delay to capture the true patient saccade. During inference, without access to the stimulus, we averaged the model's logits across sequential 1-second clips extracted from the final 7 seconds of each patient's video. The classifier was trained on 80\% of the synthetic data, validated on the remaining 20\%, and tested on the full clinical dataset. We used AUROC, AUPRC, sensitivity, specificity, and PPV metrics for evaluation.

\section{Results}
\subsection{Synthetic Waveform Validation} 
The property that all saccades must follow is the main sequence; as the amplitude of the saccade increases, the velocity also increases. In \textbf{Fig.~\ref{fig:MainSeq_syn_distribution}} below, the left plot represents the amplitude-velocity curve of a previous study, which determined that a fixed square root function was the best fitting curve for amplitude shifts less than 6 degrees and that an exponential rise function was the best fitting curve for amplitude shifts greater than 6 degrees~\cite{BehaviorResearchMethods}. The right plot of Fig.~\ref{fig:MainSeq_syn_distribution} represents the main sequence derived from our synthetic data, where we incorporated those fits into our algorithm. 
The synthetic waveforms show a main sequence that is consistent with the established main sequence in clinical settings, indicating that our synthetic saccadic waveforms are physiologically plausible.

\begin{figure}[ht]
    \centering
    \includegraphics[width=\linewidth]{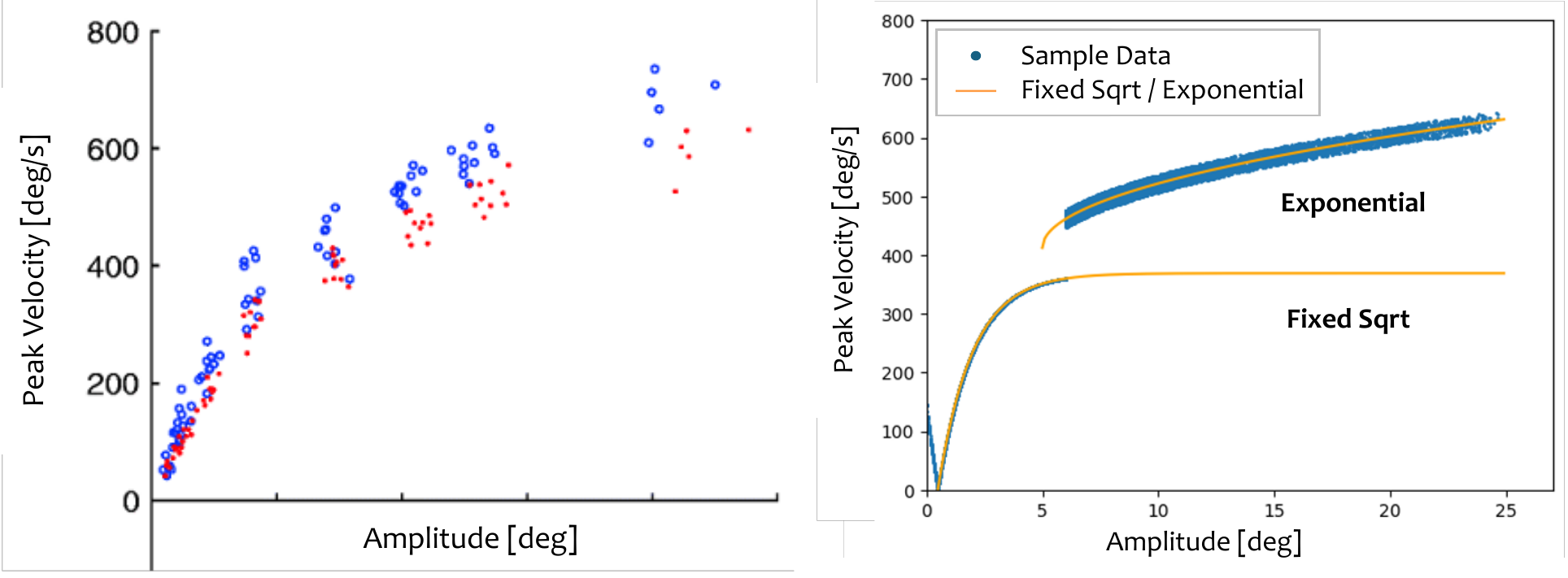}
    \caption{\textbf{Synthetic Waveform Validation.} Main sequence analysis of our synthetic waveform shows high correlation with the amplitude-velocity curve of a previous study.}
    \label{fig:MainSeq_syn_distribution}
\end{figure}

 \subsection{Synthetic Video Validation} 

After using the waveforms to guide the movement of our pupil mask and subsequently our eye videos, we re-segmented the pupil mask from the eye videos to validate whether the videos correctly followed the movement in the waveforms. In order to re-segment the pupil from our synthetic eye videos, we utilized DeepVOG~\cite{YIU2019108307} to segment the pupil mask in the generated video, fit the segmented pupil as an ellipse, and extract the video back to the waveform. The synthetic data samples achieve a $0.034\pm 0.012$ mean square error and a cosine similarity of $93.23\%$. High similarity between the synthetic waveform and the extracted waveform demonstrates the semantic preservation ability of our synthetic eye video. 

\begin{figure}[ht]
    \centering
    \includegraphics[width=\linewidth]{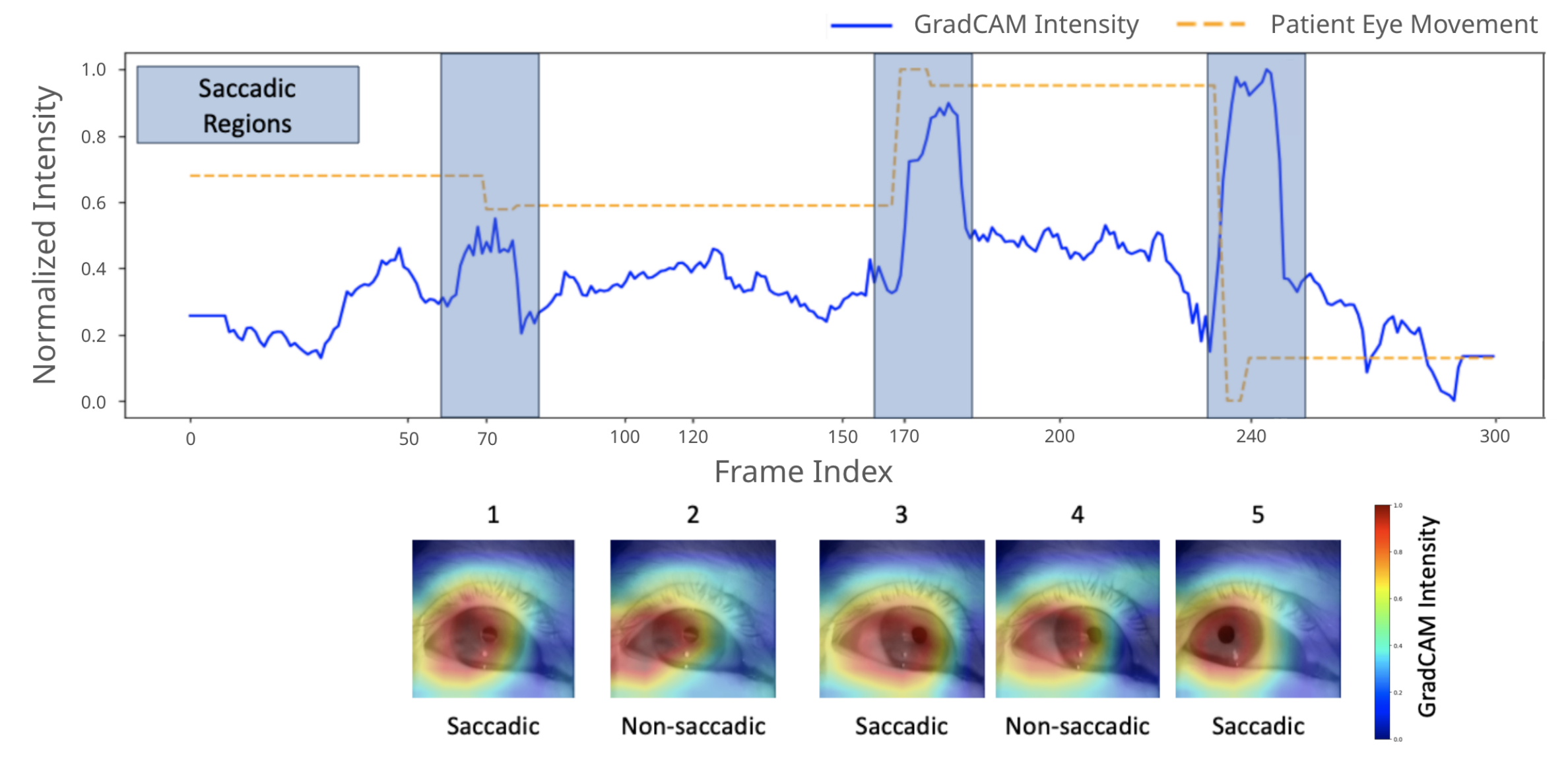}
    \caption{\textbf{Grad-CAM visualization on clinical data.} The normalized intensity of the generated eye video shows peaks during the saccadic regions. The Grad-CAM intensity map demonstrates that the model keeps focus on the eye region during eye movement.}
    \label{fig:GradCam}
\end{figure}

After testing the MViT-V2 model on clinical data, we performed Gradient-weighted Class Activation Mapping (Grad-CAM) analysis~\cite{8237336} to identify key regions of the oculographs and eye with the highest importance to model predictions. In \textbf{Fig.~\ref{fig:GradCam}}, which shows a single patient example in the clinical dataset, the Grad-CAM intensity rapidly increases when a saccade occurs. This indicates that the model focuses more heavily on the pupil during the saccadic regions compared to the non-saccadic regions. Also, as shown by the heatmap of eye images below the plot, even in the non-saccadic regions, the model consistently focuses on the pupil region throughout the whole video - mimicking current eye tracking technology and clinicians. \textcolor{black}{Furthermore, manual inspection of multiple patient examples revealed the same trend: Grad-CAM intensity peaks during saccadic regions and remains centered on the pupil, aligning with clinical expectations. The Grad-CAM visualization shown in Fig.~\ref{fig:GradCam} corresponds to a single patient example.}

\subsection{Model Performance}

Table~\ref{tab:classification_results} compares the performance of four models\textcolor{black}{: }R3D~\cite{tran2018closer}, MC3~\cite{tran2018closer}, R(1+2)D~\cite{tran2018closer}, and MViT-V2~\cite{fan2021multiscale}\textcolor{black}{, }on both synthetic and clinical datasets on distinguishing normal and abnormal saccades. The MViT-V2 model achieves the highest sensitivity and the most balanced performance across both datasets among the four model architectures.
\textcolor{black}{We report sensitivity, specificity, and positive predictive value (PPV) because they are widely used in clinical diagnostic studies. These metrics provide interpretable assessments of a model’s ability to detect neurologic abnormalities in eye movements and are highly relevant for evaluating clinical screening tools.}

\begin{table}[ht]
\centering
\caption{\textbf{Model performance on synthetic and clinical data.} MViT-V2~\cite{fan2021multiscale} achieves the best classification results on \textcolor{black}{both} synthetic data and clinical data. \textcolor{black}{Metrics are reported as binary classification (normal vs. abnormal), reflecting the intended use of the model as a screening tool to detect any saccadic abnormality. Evaluation on synthetic data was performed using a held-out subset not seen during training.}}
\caption{Model performance on synthetic and clinical data.}

\begin{tabularx}{\linewidth}{>{\centering\arraybackslash}p{3.5cm} *{6}{>{\centering\arraybackslash}X}}
\toprule
\multirow{2}{*}{\textbf{Method}} & \multicolumn{3}{c}{\textbf{Synthetic Dataset}} & \multicolumn{3}{c}{\textbf{Clinical Dataset}} \\
\cmidrule(lr){2-4} \cmidrule(lr){5-7}
 & \makecell{Sens. \\(\%)} & \makecell{Spec. \\(\%)} & \makecell{PPV \\(\%)} & \makecell{Sens. \\(\%)} & \makecell{Spec. \\(\%)} & \makecell{PPV \\(\%)} \\
\midrule
R3D~\cite{tran2018closer} & 96.5 & 95.2 & 96.8 & \textcolor{black}{51.2} & \textcolor{black}{60.4} & \textcolor{black}{48.4 }\\
MC3~\cite{tran2018closer} & 97.8 & 96.4 & 97.5 & \textcolor{black}{54.3} & \textcolor{black}{54.9 }& \textcolor{black}{52.1} \\
R(1+2)D~\cite{tran2018closer} & 98.7 & 97.9 & 97.8 & \textcolor{black}{58.1} & \textcolor{black}{56.2 }& \textcolor{black}{50.1}\\
\textbf{MViT-V2}~\cite{fan2021multiscale} & \textbf{99.3} & \textbf{99.4} & \textbf{99.8} & \textbf{\textcolor{black}{71.0}} & \textbf{\textcolor{black}{62.2}} & \textbf{\textcolor{black}{65.1}} \\
\bottomrule
\end{tabularx}
\label{tab:classification_results}
\end{table}

For classification evaluation, we grouped hypermetria and hypometria classes together as a single abnormal class. 
On synthetic data, the model achieves an AUROC of 0.99 and AUPRC of 0.98, and an AUROC of 0.76 and AUPRC of 0.67 on the clinical data. For synthetic data, the model achieves nearly perfect performance, with only 2 normal cases classified as hypermetria and 2 hypermetria cases classified as normal. However, in clinical data, the model demonstrates lower performance and struggles with accurately distinguishing normal and abnormal saccades, particularly for the hypometria class.

\section{Discussion}
By generating a fully synthetic, patient-free saccade video-oculographic dataset, we mitigate critical privacy concerns while demonstrating that models trained exclusively on synthetic data can generalize to real-world clinical applications. Our MViT-V2 model achieved an AUROC of 0.76 and a sensitivity of 71\% on out-of-distribution clinical data, showing promise as an objective digital biomarker for neurological abnormalities; especially in emergency and remote settings~\cite{wei2025teleautohints}. The observed performance gap between the synthetic and clinical datasets highlights the complexity of real-world recordings, which often contain overlapping waveform types, variable frame rates, and diverse noise artifacts~\cite{wagle2022-46}. Future work will address these limitations by expanding the synthetic pipeline to incorporate additional clinically relevant parameters (e.g., latency, velocity, conjugacy) and a broader spectrum of eye movements. Ultimately, this approach establishes a scalable foundation for developing digital biomarkers in neurology without relying on protected patient information, facilitating community-driven innovation.




%
%
\bibliographystyle{splncs04}
\bibliography{reference}

\end{document}